\documentclass[conference]{IEEEtran}

\ifCLASSINFOpdf
  \usepackage[pdftex]{graphicx}
\else
\fi
\usepackage{color,soul}
\usepackage{algorithm}
\usepackage{algpseudocode}

\usepackage{multirow}
\usepackage{xcolor}
\usepackage{todonotes}
\usepackage{stfloats}
\usepackage{graphicx}
\usepackage{multirow}
\usepackage{booktabs}
\usepackage{subcaption}

\IEEEoverridecommandlockouts

\begin{document}
%
% paper title
% Titles are generally capitalized except for words such as a, an, and, as,
% at, but, by, for, in, nor, of, on, or, the, to and up, which are usually
% not capitalized unless they are the first or last word of the title.
% Linebreaks \\ can be used within to get better formatting as desired.
% Do not put math or special symbols in the title.
\title{Adaptive Parameterization of Deep Learning Models for Federated Learning}
% Federated Learning on Non-IID Data Silos: An
% Experimental Study
%
% author names and IEEE memberships
% note positions of commas and nonbreaking spaces ( ~ ) LaTeX will not break
% a structure at a ~ so this keeps an author's name from being broken across
% two lines.
% use \thanks{} to gain access to the first footnote area
% a separate \thanks must be used for each paragraph as LaTeX2e's \thanks
% was not built to handle multiple paragraphs
%
%\author{\IEEEauthorblockN{Anonymous Authors}}
\author{Morten~From~Elvebakken,~
        Alexandros~Iosifidis,~
        and~Lukas~Esterle~% <-this % stops a space
\thanks{Morten From Elvebakken, Alexandros Iosifidis, and Lukas Esterle are with DIGIT,
the Department of Electrical and Computer Engineering, Aarhus University,
Aarhus, Midtjylland, Denmark (e-mail: mfe@ece.au.dk; ai@ece.au.dk;
lukas.esterle@ece.au.dk).
}% <-this % stops a space
}

% note the % following the last \IEEEmembership and also \thanks - 
% these prevent an unwanted space from occurring between the last author name
% and the end of the author line. i.e., if you had this:
% 
% \author{....lastname \thanks{...} \thanks{...} }
%                     ^------------^------------^----Do not want these spaces!
%
% a space would be appended to the last name and could cause every name on that
% line to be shifted left slightly. This is one of those "LaTeX things". For
% instance, "\textbf{A} \textbf{B}" will typeset as "A B" not "AB". To get
% "AB" then you have to do: "\textbf{A}\textbf{B}"
% \thanks is no different in this regard, so shield the last } of each \thanks
% that ends a line with a % and do not let a space in before the next \thanks.
% Spaces after \IEEEmembership other than the last one are OK (and needed) as
% you are supposed to have spaces between the names. For what it is worth,
% this is a minor point as most people would not even notice if the said evil
% space somehow managed to creep in.

% The paper headers
\markboth{Paper, October~2022..}%
{Shell \MakeLowercase{\textit{et al.}}: Adaptive parameterization of deep learning models for Federated Learning }
% The only time the second header will appear is for the odd numbered pages
% after the title page when using the twoside option.
% 
% *** Note that you probably will NOT want to include the author's ***
% *** name in the headers of peer review papers.                   ***
% You can use \ifCLASSOPTIONpeerreview for conditional compilation here if
% you desire.

% If you want to put a publisher's ID mark on the page you can do it like
% this:
%\IEEEpubid{0000--0000/00\$00.00~\copyright~2015 IEEE}
% Remember, if you use this you must call \IEEEpubidadjcol in the second
% column for its text to clear the IEEEpubid mark.

% use for special paper notices
%\IEEEspecialpapernotice{(Invited Paper)}

% make the title area
\maketitle

% As a general rule, do not put math, special symbols or citations
% in the abstract or keywords.
\begin{abstract}
Federated Learning offers a way to train deep neural networks in a distributed fashion. While this addresses limitations related to distributed data, it incurs a communication overhead as the model parameters or gradients need to be exchanged regularly during training. This can be an issue with large scale distribution of learning tasks and negate the benefit of the respective resource distribution. In this paper, we we propose to utilise parallel Adapters for Federated Learning. Using various datasets, we show that Adapters can be incorporated to different Federated Learning techniques. We highlight that our approach can achieve similar inference performance compared to training the full model while reducing the communication overhead by roughly 90\%. We further explore the applicability of Adapters in cross-silo and cross-device settings, as well as different non-IID data distributions. 
\end{abstract}

% Note that keywords are not normally used for peerreview papers.

% For peer review papers, you can put extra information on the cover
% page as needed:
% \ifCLASSOPTIONpeerreview
% \begin{center} \bfseries EDICS Category: 3-BBND \end{center}
% \fi
%
% For peerreview papers, this IEEEtran command inserts a page break and
% creates the second title. It will be ignored for other modes.
\IEEEpeerreviewmaketitle

\section{Introduction}
Machine learning generates models for specific problems by resolving an optimization problem. This is usually achieved by a single system having access to a large dataset. With increasing amounts of data available, the effectiveness of machine learning methods improves as well. However, at the same time, requirements for computational power in a single computing infrastructure are rising alongside~\cite{Bianco2018benchmark}.
Emerging privacy concerns and rising data regulations put additional burdens on centralised data collection and make it more problematic.
%In addition, privacy concerns emerge and data regulations increase. Therefore collecting the dataset in a single entity becomes more problematic. 
Hence, training machine learning models in a centralized setting faces many challenges making the solution less feasible, leading to other solutions entering the field. One of these solutions is to train the deep learning model through federated learning (FL) \cite{fedavg}. 

\begin{figure}[]
   \centering
   \includegraphics[width=0.95\linewidth]{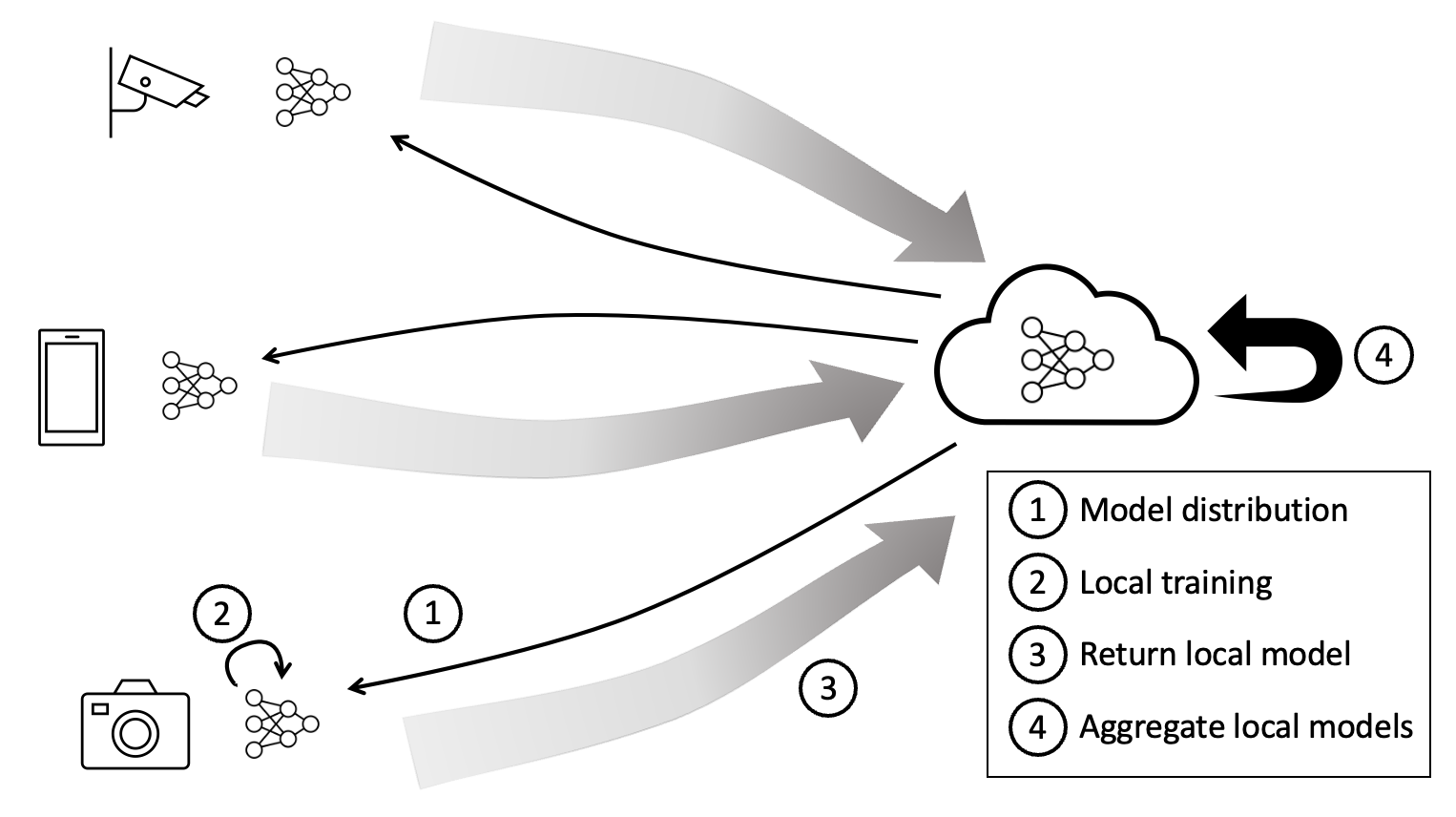}
    \caption{Federated Learning cycle: (1) individual clients receive a model and (2) train on their own data. (3) The clients share the updated model with a server (possibly in the cloud). (4) The central server performs combines the received updated models and (1) distributes them back to the clients. These steps constitute a communication round, which can be repeated to further update the model. Not all clients need to receive and update the model at each communication round.}
    \label{fig:FLIllustration}
\end{figure}

FL can be used in different scenarios ranging from edge devices~\cite{edgeDevices} and smartphones~\cite{SmartphoneFL} to healthcare informatics~\cite{healthcareinformatics}. The core idea in FL is decentralized machine learning where one trains a model at several client (edge) devices and aggregates the resulting locally trained models. This means no exchange of data between devices. Most FL algorithms rely on an iterative process of decentralized learning, which is illustrated in Figure \ref{fig:FLIllustration}. In each communication round, participating clients receive a global model, train it on local data and send it back to a coordinating server. The created federated model, integrating the shared individual models, is distributed among the participants at the end of the communication round.

Iteratively exchanging the global model as well as the learned local models leads to large amounts of data transmitted between the server and clients. To overcome this issue, we propose using Adapters for convolutional neural networks. Adapters have originally been proposed as an alternative to fine-tuning a model in multi domain learning~\cite{AdaptersDomainSrebuffi, AdaptersDomain}. 
Using a so-called domain-agnostic base model, Adapters are added in either a series or a parallel configuration. This reduces the number of parameters that need training for a new domain to only the Adapters and other domain specific layers such as batch normalization. 
The clients then learn a set of Adapters collectively through FL. In communication rounds, only these Adapters are shared, effectively reducing the amount of exchanged data.

It is commonly expected in FL that the set of clients has to deal with data which is not independent and identically distributed (non-IID) among all clients%. This non-IID data can have different shapes, namely (1) Label distribution skew, (2) Feature distribution skew, (3) same labels but different features, (4) same features but different labels and (5) quantity skew
~\cite{FLsurveyAdvancesAndOpenProblems}. In this work, we consider label distribution skew and feature distribution skew to simulate non-IID data. With label distribution skew, the label distribution $P(y_{i})$ varies across the clients. This simulates situations appearing in several applications, such as hospitals specializing in certain areas. With feature distribution skew, the feature distributions $P(x_{i}) = y_i$ differ from client to client for the same label. An example could be an image of a dog and a drawing of a dog, that are both to be classified as $y_{i} = dog$. An example of label distribution skew can be seen on Figure \ref{fig:dirichlet} where the clients have varying amount of data for each label. We focus on horizontal FL, where the clients have different samples but they all share the same feature space.

We focus on two scenarios of FL, namely cross-device and cross-silo FL~\cite{FLsurveyAdvancesAndOpenProblems}. In cross-device FL, data is split among the participating clients. Here, FL is usually performed on a large number of embedded devices with limited resources and computational power. For this, smaller models are employed to allow for suitable training on the device. In cross-device settings, it is common %to expect 
that only a fraction of clients report back during communication rounds.
In contrast, cross-silo FL has less resource % and computational 
restrictions and uses fewer clients. Here, each client has access to more data and more processing power. Cross-silo FL is usually considered without client instability and failure to report back to the server, allowing all clients to participate in each communication round. The number of clients in each setting varies. The authors of \cite{FLsurveyAdvancesAndOpenProblems} use between 2-100 in cross-silo and up to $10^{10}$ in cross-device. 

In summary, our contributions are: (i) We propose a novel approach to FL, reducing the amount of data to be transmitted in each FL communication round by introducing Adapters to FL. %We propose the use of Adapters in conjunction with federated learning.
(ii) We explore the trade-off between the amount of data exchanged against the inference accuracy. (iii) We examine the application of Adapters for FL in both cross-silo and cross-device settings across varying non-IID settings. 

In addition, the proposed approach has two more benefits. First, it increases security as only Adapters are exchanged among the clients in each communication round. Even if the communication is intercepted, an attacker does not necessarily have access to a functioning model, as Adapters are useless without the base model. Second, our approach can be used as an add-on to existing FL algorithms. 

% The remainder of the paper is structured as follows. In the next section we elaborate on related and prior work. In Section~\ref{sec:method} we present the proposed approach utilising Adapters for FL. Section~\ref{sec:experiments} outlines the experimental protocol before we discuss the results in Section~\ref{sec:results}. We conclude the paper with the main findings and an outline on future work in Section~\ref{sec:conclusion}.

%\begin{figure*}[!]
%	\centering
%	\includegraphics[height=1\linewidth,angle=90]{images/Fig_2/ArchitectureResnet26.pdf}%ArchResnet26.png}
%	\caption{Parallel Adapters on the ResNet26 architecture. The weights of the base model (in grey) are frozen, and only Adapters (in green) are trained.}
%	\label{fig:adaptersResnet}
%\end{figure*}

%\begin{figure}[!h]
%   \centering
%   \includegraphics[width=0.5\linewidth]{images/Fig_3/adaptersBlock.pdf}
%    \caption{Configuration of parallel Adapters for two consecutive blocks in ResNet26. $W_{l}$ parameters are frozen.}
%    \label{fig:adaptersBlock}
%\end{figure}
\section{Related work}
\label{sec:relwork}
FedAvg~\cite{fedavg} is the baseline FL algorithm for most research in this area. FedAvg trains a shared model following an iterative process formed by four steps with a coordinating server and involved clients with local data. First, the server distributes the global model to selected clients. Each of them then trains the model on its local data. The server aggregates the locally updated models and takes a weighted average of the received parameters to form the global model for the next step in the iterative algorithm. With this approach, FL enables multiple parties to collaboratively train a model, without exchanging local data.
FL has expanded since the emergence of FedAvg, examining various aspects of the field such as communication costs, data privacy, and performance under different non-IID settings.

Under non-IID settings in FL, each client participating in the training moves towards its own optima. Under heavy non-IID data distributions this leads to what is called \textit{client drift}~\cite{scaffold}. Research how to handle client drift focuses on non-IID data settings where devices included in the FL algorithm will diverge towards their local optima instead of the optima of the global problem. In this setting it is harder, if not impossible, to train models that reach near the optima of the global problem purely based on FL. Algorithms such as FedProx~\cite{fedprox} and FedNova~\cite{fednova} seek to better handle the heterogeneity of the clients. FedProx locally calculates a proximal term to restrict local updates to be closer to the initial globally distributed model, thereby limiting updates that diverge far from the starting point. The proximal term needs to be tuned to the problem at hand. A too small proximal term means the proximal term has almost no effect, whilst a too large proximal term leads to small limited updates which impacts convergence speed. FedNova aims to improve FedAvg in the aggregation stage by accounting for different amounts of local steps performed on each participating client. This could be due to variations in available computation power, time constraints or amount of local data. To counter global updates being biased towards clients that perform more local updates, FedNova uses a normalized averaging method. 
{Our presented approach allows to extend these and other FL methods and reduces the respective communication effort.}

%Other papers that go into AFL and forth
There is a lot of different research into reducing communication in FL. Some aim to learn the global model through a single communication round, this is referred to as one-shot FL~\cite{oneshotfl, fedkt}. Others approach the communication problem by aiming to jointly improve convergence time and training loss, further they employ quantization to reduce the payload~\cite{chen2021communication}. Quantization and other compression methods are used in some previous works~\cite{sattler2019robust, fedzip, li2021communication, aoq}. Fast-Convergent Federated Learning (FOLB) performs intelligent sampling of clients in each round, optimizing the convergence speed~\cite{nguyen2020fast}.
{These techniques still require additional processing or implementation in order to reduce the amount of exchanged information. This leads to a trade-off between training, compression, and data exchange. %present an opportunity to improve the presented approach using Adapters and further reduce communication effort. 
}

There are other papers examining expensive communication in FL by exchanging fewer parameters as well. Adaptive Parameter Freezing (APF) adaptively freezes parameters that stabilize during training and no longer exchanges these during FL rounds in intermittent periods~\cite{AdaptiveParFreezing}. The periods are to allow parameters that only temporarily stabilize to be updated later in the process. With APF they can reduce communication by 60\% during training. Federated Parameter Efficient Fine-Tuning (FedPEFT) investigates the use of parameter-efficient finetuning for large pre-trained models~\cite{fedpeft}. One of the methods they investigate is similar to ours with a type of Adapter injected, but for Transformers~\cite{adapterTransformers}. 
Another approach called parameter Prediction Based FL (PBFL) takes advantage of predicting values to aggregate the model, reducing the required communication rounds, stating an improved communication efficiency by more than 66\%~\cite{li2023pbfl}.
For neural machine translation (NMT) models in FL a new layer called ``Controller'' layer is proposed~\cite{comEfFLNM}. They use the same approach of only exchanging certain layers from the full model and freezing the base model.
%{\color{red}In contrast to these approaches, we are not freezing any parameters but utilise Adapters for refining a domain-agnostic base model. This allows us to reduce communication effort by about 90\%.}

In contrast to these state-of-the-art approaches in FL, we are adapting deep neural networks using residual Adapters as applied in multi-domain learning~\cite{rebuffiFirstAdapters}. Instead of fine-tuning a network to different domains, the use of a small set of residual Adapters for each new domain is proposed, keeping the base model the same for each domain~\cite{AdaptersDomain, AdaptersDomainSrebuffi}. 
%Adapting deep neural networks using residual Adapters is from multi-domain learning~\cite{rebuffiFirstAdapters}. Instead of fine-tuning a network to different domains, the use of a small set of residual Adapters for each new domain is proposed, keeping the base model the same for each domain~\cite{AdaptersDomain, AdaptersDomainSrebuffi}. 
{We show that resulting models using adaptive parameterization in FL for training perform almost on par with the finetuned full base model whilst lowering the amount of parameters to be trained and exchanged in each communication round.}
%The resulting models perform almost on par with models obtained by finetuning the base model whilst lowering the amount of parameters to be trained for each new domain. 
%
%The residual Adapters are convolutional filters of kernel size 1$\times$1.
%Each convolutional layer in the ResNet26 configuration has its own residual Adapter, either in a series or a parallel configuration. We only examine the parallel configuration in this paper. This approach is simpler in configuration and needs no extra batch normalization layer when adapting the model. When the kernels of the convolutional layers that are adapted are of size 3$\times$3, the domain-specific Adapters are roughly 9 times less than the base model parameters, since we still need domain-specific layers such as batch normalization~\cite{AdaptersDomain}.
%
%It is on this assertion that we examine the use of Adapters to train a model within FL to lower communication. 
By exchanging Adapters of kernel size $1 \times 1$ instead of the base model convolutional layers of size $3 \times 3$, each communication round of FL is roughly 9 times smaller, since we still have bias and batch normalization. The base model needs to be distributed as well the first time, unless this is known at the client side already. This could be the case if a client has data from multiple domains (multi-domain learning), and requires a model to handle each of them.

\section{Adapters for federated learning}
\label{sec:method}
%\subsection{}
In this work, we investigate the use of Adapters for deep neural networks within FL. Our primary objective is to learn a shared global model to perform inference tasks on individual devices using deep neural networks. The implementation of Adapters follows~\cite{AdaptersDomain,AdaptersDomainSrebuffi}, adding parallel Adapters to an Imagenet pre-trained ResNet26~\cite{resnet}. We freeze the convolutional layers of the pretrained model, which we denote as the base model. Training a full model requires all parameters to be transmitted in each communication round, whereas when we use Adapters we only need to transmit parameters that are not specific to the base model.

A convolutional neural network is built from several different building blocks, where the main block is the convolutional layer. The model learns a set of parameters $W$ that represents learned features to achieve a goal. We denote the set of convolutional layers as $L$, where $W_{l}\in W$ represents the parameters of convolutional layer $l$. 
For each convolutional layer with parameters $W_l$ of the size $c \times c \times I_l$, we inject an Adapter $a_{l}$ of size $1 \times 1 \times I_l$. It is important to note that each Adapter $a_{l}, \:l=1,\dots,L$ will have the same output size as its corresponding base model layer with parameters $W_{l}$, since both the base layer and the Adapter have the same number of filters $I_l$, and appropriate padding is applied when the input is processed by the parameters of the layer $W_{l}$. 
\begin{figure*}[h!]
\centering
\begin{subfigure}[b]{0.41\textwidth}
    \includegraphics[width=\textwidth]{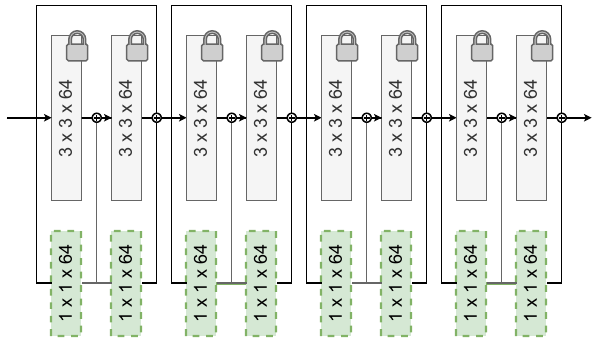}
    \caption{Resnet26 architecture}
    \label{fig:adaptersResnet}
\end{subfigure}
\hfill
\begin{subfigure}[b]{0.58\textwidth}
    \includegraphics[width=\textwidth]{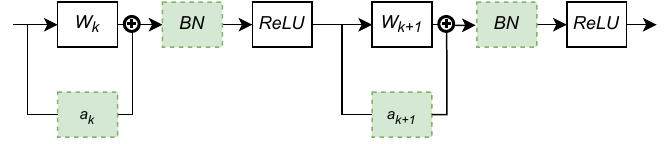}
    \caption{Single residual block}
    \label{fig:adaptersBlock}
\end{subfigure}
\caption{a) Parallel Adapters configuration on the ResNet26 architecture early residual blocks. The weights of the base model (in grey) are frozen, and only Adapters (in green) are trained. b) Closer look at one residual block}
\end{figure*}

Figure \ref{fig:adaptersBlock} shows a configuration of parallel Adapters for two consecutive blocks in the ResNet26 architecture. We denote the convolutional layer of each block $\mathcal{L}_{l}$ as $\mathcal{L}_{l}(x;W_{l}) = g(W_{l}\ast x)$ where $g(\cdot)$ is the Rectified Linear Unit (ReLU) activation function and $\ast$ is the convolution operation. With Adapters injected, the $l$-th convolutional layer in Figure \ref{fig:adaptersBlock} can be denoted as:
\begin{equation}
    \mathcal{L}_{l,a}(x;W_{l},a_{l}) = g(W_{l}\ast x + a_{l}\ast x). \label{Eq:ConvLayerAdapters}
\end{equation}
The batch normalization layer with domain-specific parameters $\mu_{l}, \sigma_{l}$ is applied on $(W_{l}\ast x + a_{l}\ast x)$ before applying the activation function $g(\cdot)$.

Using the definition of the convolutional layer in Eq. \ref{Eq:ConvLayerAdapters}, the domain-specific parameters that we need to train in each communication round are contained in the Adapters $a_{l}, \:l=1,\dots,L$, the batch normalization parameters $\mu_{l}, \sigma_{l}, \:l=1,\dots,L$, and the weights of the fully-connected output layers. 
The ResNet26 architecture forms the base model with parameters $W$, which we can parametrize using residual Adapters, as seen on Figure \ref{fig:adaptersResnet} \cite{AdaptersDomainSrebuffi,  AdaptersDomain}. ResNets consists of residual blocks of two consecutive convolutional layers of $c \times c \times I_l$, where the number of base model parameters of said residual block can be expressed as $2(c^{2}I_l^{2}+I_l)$. The 2 accounts for the two convolutional layers and the additional $I_l$ contains biases of the convolutional layer. Since the ResNet26 does not apply bias in its convolutional layer, the base model parameters per block can be expressed as $2(c^{2}I_l^{2})$. We can then express the domain-specific parameters corresponding to that residual block. With an Adapter of size $1 \times 1 \times I_l$, the domain specific parameters can be expressed as $2(I_l^{2}+2I_l)$. The $2I_l$ here accounts for the batch normalization parameters, which are not domain specific. 
Since the ResNet26 uses $3 \times 3$ convolutional layers and the Adapters are $1 \times 1$ convolutional layers, we can effectively lower the data exchange by a factor of approximately 9. 
%the amount of parameters that needs to be transmitted by a factor of approximately 9.

%It has been shown that the higher number of classes used when pre-training the network, the better it is at adapting to new tasks~\cite{AdaptersDomainSrebuffi}. We use the same base model used in previous work into residual Adapters. When training the Adapters, the base model's parameters $W_{l}$ are frozen, meaning they are not finetuned during training.

We distribute the data across a set of clients $\mathcal{C} = \{c_1, c_2, ..., c_i,... c_n\}$ participating in the training of the model. Importantly, $\forall c_i \in \mathcal{C}$ have the same pre-trained base model and the same task. For the simulated cross-device experiments, we use random sampling with replacement to pick a fraction $f$ from the total population of the clients in $\mathcal{C}$ that participate in each training round. The participating clients in a communication round is denoted as $\mathcal{C}_{f}$. 
\section{Experiments}
\label{sec:experiments}

To investigate the effectiveness of using Adapters in FL we conducted experiments across multiple datasets, data distributions, and FL algorithms. 

We used three datasets, namely SVHN~\cite{svhn}, CIFAR10~\cite{cifar10} and FashionMNIST~\cite{fmnist}. We resized all images of the datasets to $72 \times 72$ pixels, following the implementations in~\cite{AdaptersDomainSrebuffi, AdaptersDomain}. The data is further divided into different non-IID distributions. Specifically, we experimented with different settings of non-IID data in both a cross-silo and cross-device settings. To simulate the different non-IID data distributions, we followed the implementation in~\cite{DirichletFLSilo}. We simulated two different label-based imbalances, quantity-based label imbalance and distribution-based label imbalance, as well as a feature-based distribution skew.

The first type of label imbalance is quantity-based label imbalance, where all clients have data samples of a specified number of labels. That is, each client has samples from the full experimental dataset containing $M$ classes. Different settings of $M$ lead to different distributions. The setting $M = 1$, for example, refers to each client only having samples belonging to a single class. Meaning that if there are 10 clients and 10 classes, each class is only found on one client. We examine the ranges with $|M| = [1, 3]$.

\begin{figure}[H] %H
   \centering
   \includegraphics[width=1\linewidth]{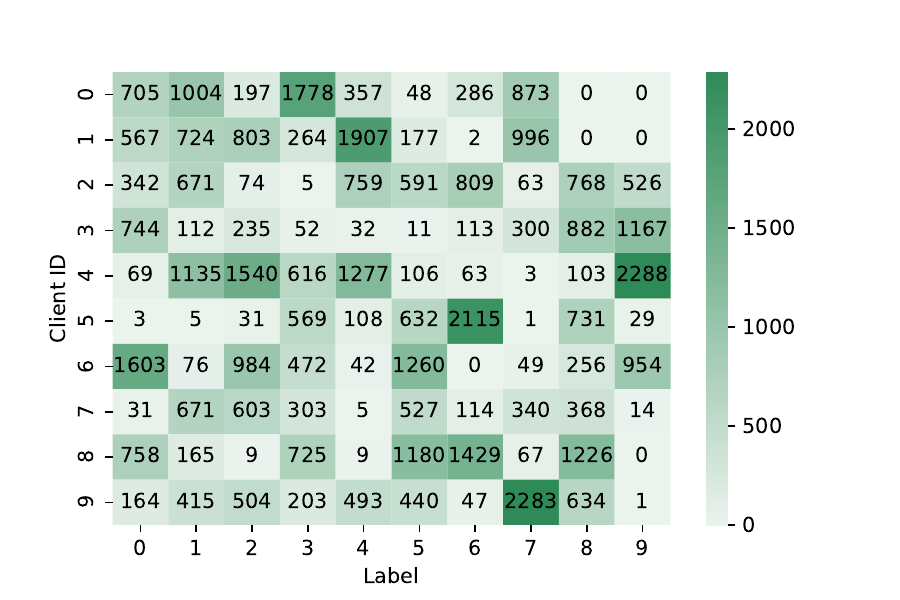}
    \caption{Distribution of Cifar10 across 10 clients for distribution-based label imbalance.}
    \label{fig:dirichlet}
\end{figure}

The second type of label imbalance is distribution-based label imbalance, which uses a Dirichlet distribution to divide the dataset. Each client has data based on a proportion of the samples of each label according to a Dirichlet distribution. We sample $p_{k} \sim Dir_{N}(\beta)$ and allocate $p_{k,j}$ proportion of class k to client j. $\beta$ is the concentration parameter, the closer to zero the more unbalanced the distribution. We use a $\beta = 0.5$. An example distribution is shown for Cifar10 for the cross-silo scenario with $ Dir(0.5)$ in Figure~\ref{fig:dirichlet}.

The noise-based feature distribution skew first divides the dataset homogeneously across clients. A different level of Gaussian noise is then added to each client's local dataset to obtain the different feature distributions. A defined noise level $\sigma$ is added based on $\hat{x} \sim Gau(\sigma * i/C)$ for client $c_{i}$. The Gaussian distribution $Gau(\sigma * i/C)$ has mean 0 and variance $\sigma * i/C$. We use a noise level of $\sigma = 0.1$ and this is represented as $Gau(0.1)$ hereafter. 

\begin{figure*}[h!]
\centering
\begin{subfigure}{0.3\textwidth}
    \includegraphics[width=\textwidth]{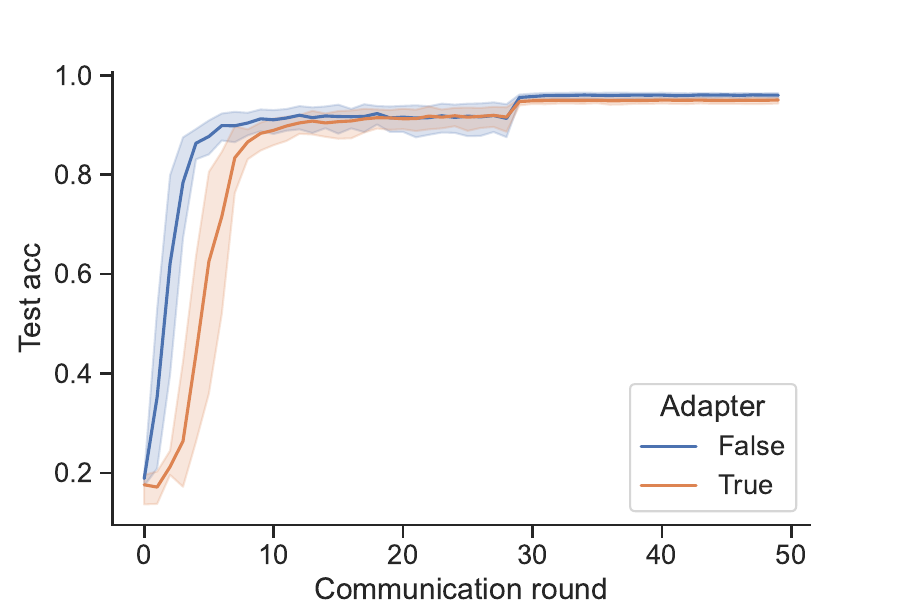}
    \caption{FedAvg SVHN}
    \label{fig:first}
\end{subfigure}
\hfill
\begin{subfigure}{0.3\textwidth}
    \includegraphics[width=\textwidth]{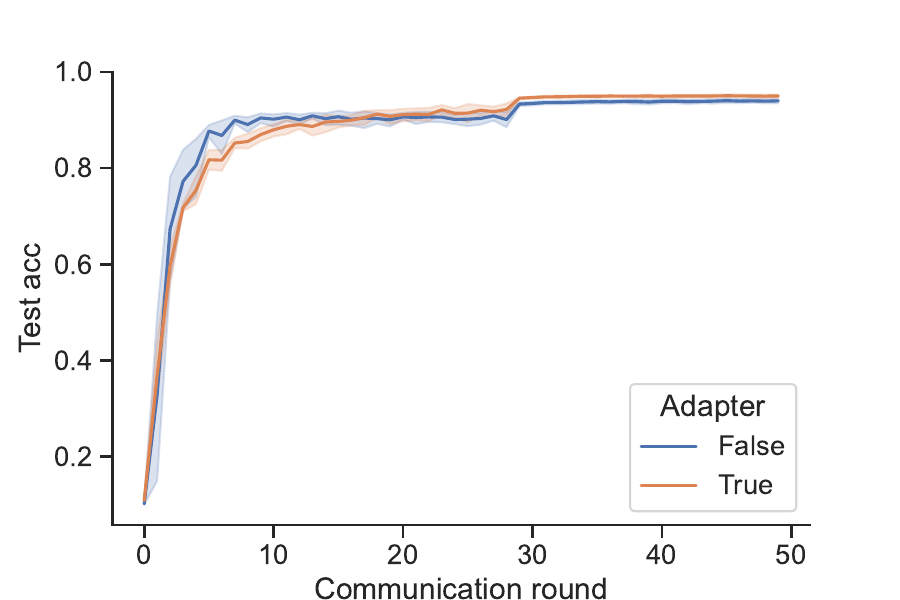}
    \caption{FedAvg Cifar10}
    \label{fig:second}
\end{subfigure}
\hfill
\begin{subfigure}{0.3\textwidth}
    \includegraphics[width=\textwidth]{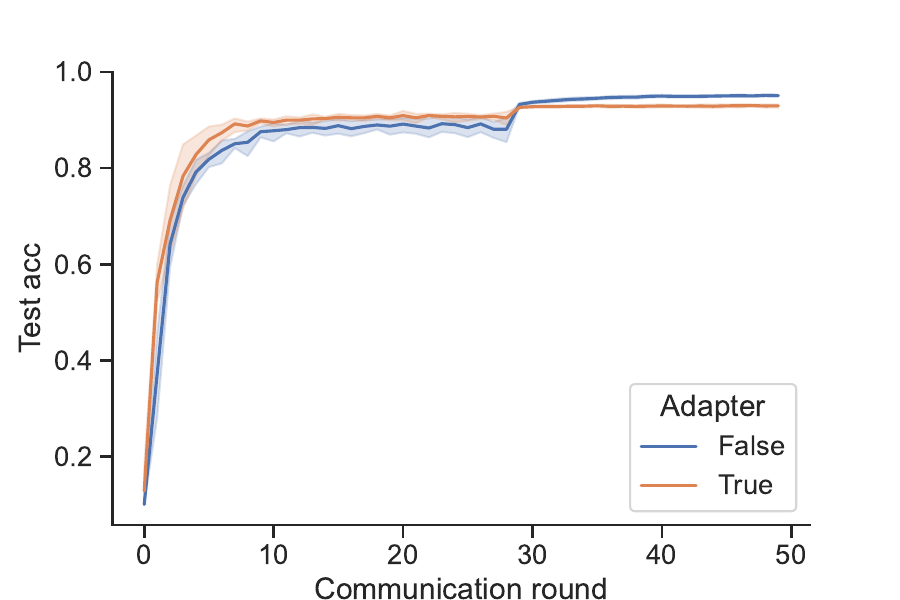}
    \caption{FedAvg FashionMNIST}
    \label{fig:third}
\end{subfigure}
\begin{subfigure}{0.3\textwidth}
    \includegraphics[width=\textwidth]{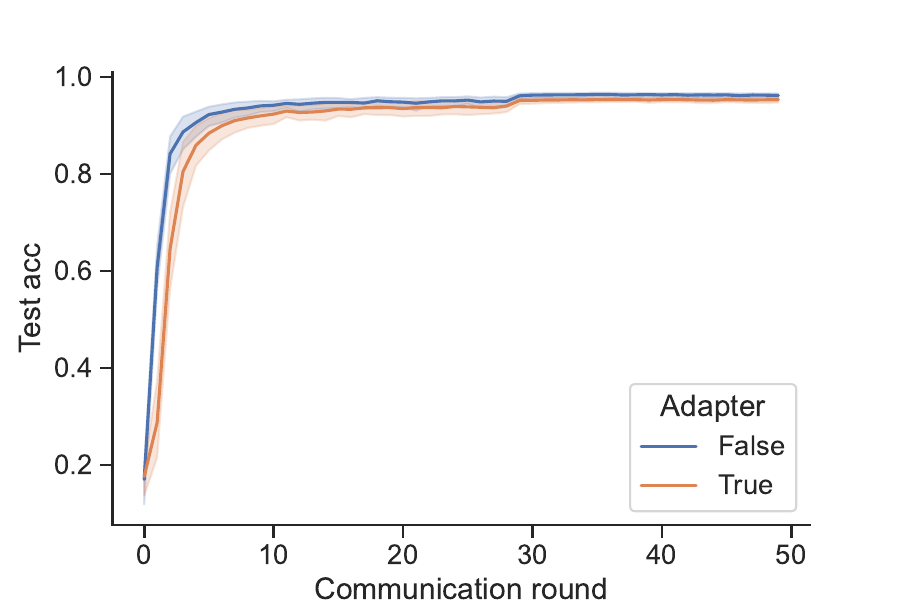}
    \caption{FedProx SVHN}
    \label{fig:first}
\end{subfigure}
\hfill
\begin{subfigure}{0.3\textwidth}
    \includegraphics[width=\textwidth]{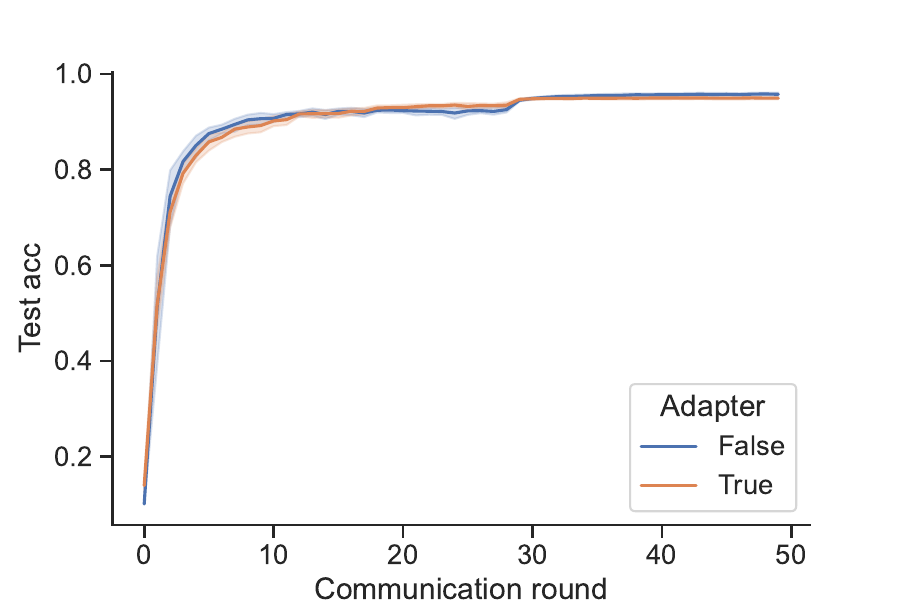}
    \caption{FedProx Cifar10}
    \label{fig:second}
\end{subfigure}
\hfill
\begin{subfigure}{0.3\textwidth}
    \includegraphics[width=\textwidth]{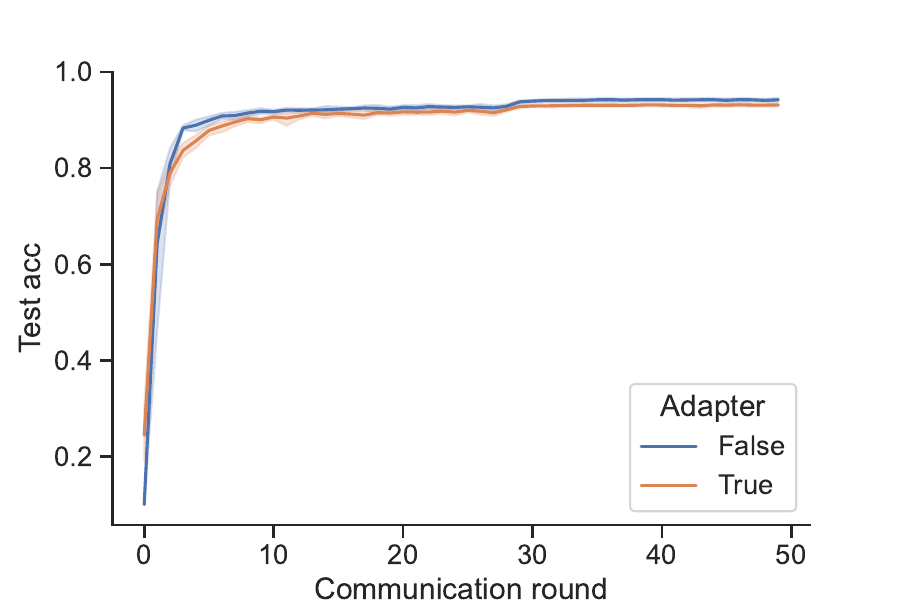}
    \caption{FedProx FashionMNIST}
    \label{fig:third}
\end{subfigure}
\begin{subfigure}{0.3\textwidth}
    \includegraphics[width=\textwidth]{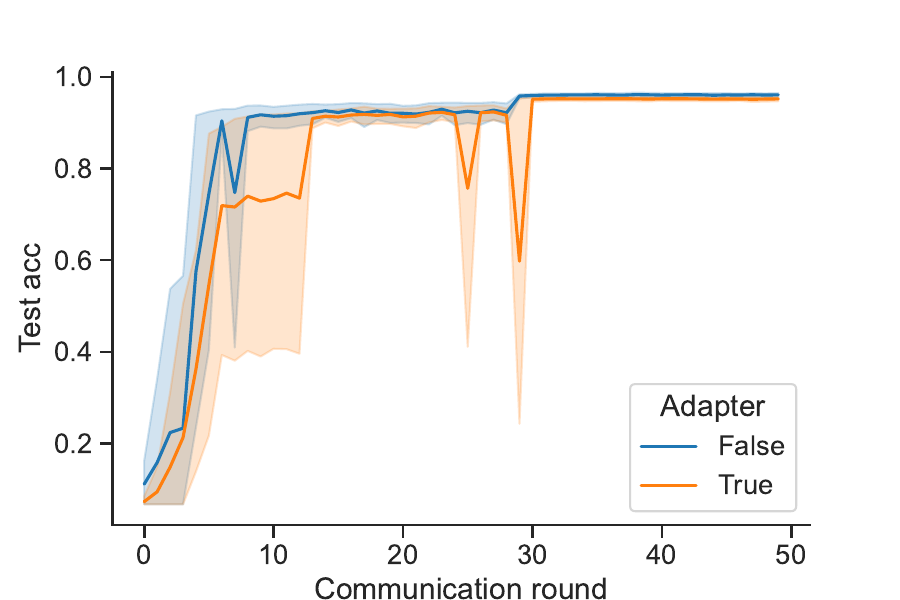}
    \caption{FedNova SVHN}
    \label{fig:first}
\end{subfigure}
\hfill
\begin{subfigure}{0.3\textwidth}
    \includegraphics[width=\textwidth]{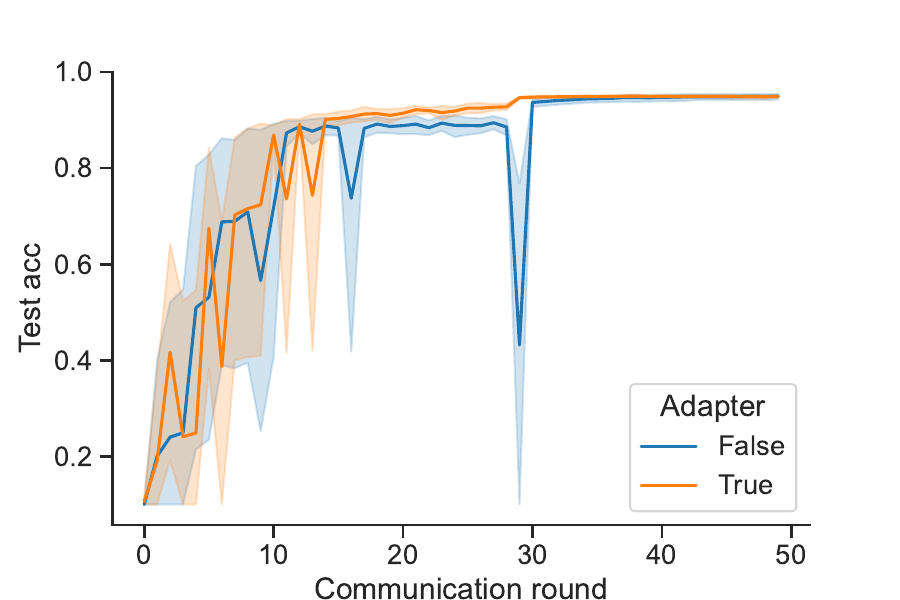}
    \caption{FedNova Cifar10}
    \label{fig:second}
\end{subfigure}
\hfill
\begin{subfigure}{0.3\textwidth}
    \includegraphics[width=\textwidth]{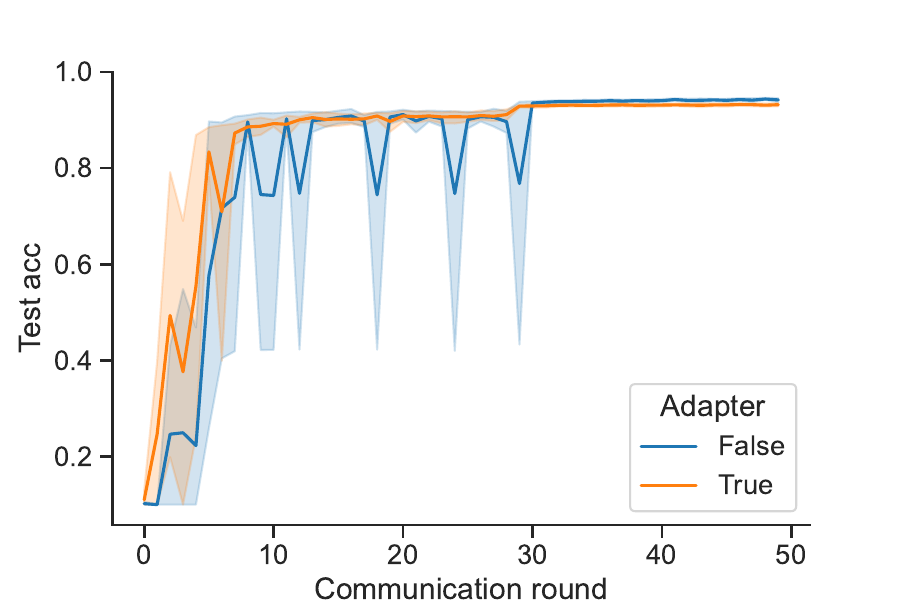}
    \caption{FedNova FashionMNIST}
    \label{fig:third}
\end{subfigure}
        \vspace{5pt}
\caption{Performance (accuracy) on the test set after each communication round in FedAvg, FedProx and FedNova. We show results for SVHN, Cifar10 and FashionMNIST, comparing the model with and without Adapters}
\label{fig:fedAlgsCrossSilo}

\end{figure*}

% \todo[inline]{this non-iid information should go to the experiments.}

\vspace{-20pt}
\subsection{Results}\label{sec:results}
Each FL algorithm is trained for 50 communication rounds using a batch size of 64. All results show the top 1 accuracy. Each experiment is run for five different random seeds and the mean and standard deviation of these runs are shown on Table ~\ref{tab:fullTableResults}. The last column shows the average difference in accuracy between finetuning and adapters. The test accuracy after each communication round for cross-silo in the distribution based label imbalance can be seen on Figure \ref{fig:fedAlgsCrossSilo}.
We use SGD as local optimizer with a weight decay of $0.0005$ and momentum of $0.9$. For distribution based label imbalance, feature based noise imbalance, and homogeneous partition, we use a learning rate (LR) of $0.1$ and decrease this to $0.01$ at round 30. For quantity-based label imbalance we use a lower starting LR of $0.01$ and decrease this at round 30 to $0.001$. This is due to instability during training, which is discussed later in the paper. The weight of the proximal term $\mu$ in FedProx is tuned from $\{0.1, 0.01, 0.001\}$ and set at $0.01$ across experiments unless otherwise specified. 

\textbf{Cross-silo setting} in our paper has 10 clients and all clients participate in each round of federated learning as done in the previous study of cross-silo FL we follow~\cite{DirichletFLSilo}.
Each client trains locally for $E = 1$ epoch, before the server aggregates the local models and transmits the aggregated parameters. We explored different values for $E = \{1,2,3,5,10\}$. In cross-silo settings, as simulated here, each client still has a substantial amount of data, and we found that increasing local computation leads to little change in performance, whilst a higher $E$ actually degraded performance.

For quantity-based label imbalance we see that finetuning the entire network using FL performs only slightly better on average compared to training Adapters alone. The standard deviation over the five runs  %of different random seeds used to distribute the data,
shows a lower or comparable value for Adapters, except in the case of FashionMNIST with FedProx.

\textbf{Cross-device setting} investigates the same non-IID strategies as before, but divides the data among 100 clients. Cross-device considers up to $10^{10}$ clients, and  therefore could have simulated a much higher number of clients. However it is also the case that ResNets can be out of the typical range of options to be deployed at smaller edge devices that would range in these environments. We set the number of clients to the lower end of what is expected in a cross-device setting, but still adhere to properties of cross-device such as only a fraction of clients reporting in each round. For future studies the use of higher client numbers may be interesting.
We use a fraction of $f = 0.2$ clients participating in each communication round across all experiments in this setting. A lower fraction than $0.2$ impacted the stability of the training, resulting in lower accuracy and convergence ability of the FL algorithms. Preliminary tests revealed that $E = 5$ epochs of local calculation produces before the aggregation stage results in the best performance with $f = 0.2$, again, tuned  from $\{1,2,3,4,5,10\}$.  
%Each client performs $E = 5$ local epochs before the aggregation stage, tuned from $\{1,2,3,4,5,10\}$.
In this setting each client has roughly a tenth of the data compared to the cross-silo setting. 
%Preliminary tests revealed that 5 epochs of local calculation produces the best performance with $f = 0.2$. 
We speculate that the higher 
%preferable 
$E$ in cross-device compared to cross-silo may be because each client has less data. Furthermore, as only a fraction of clients reports in, the experiments means each rounds aggregated model has trained on less unique data than in cross-silo. 
%
%In the cross-device experiments we found that the lower fraction $f$ of clients reporting in, the higher $\mu$ is preferable. With a fraction of $0.2$ we found that $0.01$ and $0.1$ resulted in similar accuracy, where $0.1$ took longer to converge. Lowering the fraction $f$ to 0.1 of clients reporting in each round we found that FedProx with $\mu = 0.1$ was best. With fewer clients reporting in, the likelihood of those clients being biased towards a specific sub-set of classes is increased. I.e., clients in that communication round may be heavily skewed towards certain samples. For larger amounts of local computations, a lower value of $\mu$ performs the best.
%
% Please add the following required packages to your document preamble:
% \usepackage{multirow}
% \usepackage{graphicx}

	\noindent
\begin{table*}
	\centering
\resizebox{.90\textwidth}{!}{
\begin{tabular}{{lc cc cc cc c }}  
	\toprule[1.3pt]
	\multicolumn{2}{l}{\textbf{Cross-Silo}}                                        & \multicolumn{2}{c}{FedAvg}                                       & \multicolumn{2}{c}{FedProx}                                     & \multicolumn{2}{c}{FedNova}                                      &   \\ 
	
	\cmidrule(r){1-2} \cmidrule(r){3-4} \cmidrule(r){5-6} \cmidrule(r){7-8}
	
	\multicolumn{1}{l}{Dataset}                       & Partitioning      & \multicolumn{1}{l}{w/o adapter} & \multicolumn{1}{l}{w/ adapter} & \multicolumn{1}{l}{w/o adapter} & \multicolumn{1}{l}{w/adapter} & \multicolumn{1}{l}{w/o adapter} & \multicolumn{1}{l}{w/ adapter} & \multicolumn{1}{r}{Average difference}                  \\
	
	\cmidrule(r){1-1} \cmidrule(r){2-2} \cmidrule(r){3-3} \cmidrule(r){4-4} \cmidrule(r){5-5} \cmidrule(r){6-6} \cmidrule(r){7-7} \cmidrule(r){8-8} \cmidrule(){9-9}
	
	\multicolumn{1}{l}{\multirow{5}{*}{SVHN}}         & \textit{Dir(0.5)} & 96.00 $\pm$ 0.63                & 95.03 $\pm$ 0.80                & 96.15 $\pm$ 0.85                & 95.27 $\pm$ 0.69               & 96.07 $\pm$ 0.63                & 95.19 $\pm$ 0.58                & 0.91                                    \\
	\multicolumn{1}{c}{}                              & \#M = 1           & 19.57 $\pm$ 0.16                & 19.14 $\pm$ 0.33                & 19.66 $\pm$ 0.26                & 19.33 $\pm$ 0.32               & 8.81 $\pm$ 0.79                 & 9.00 $\pm$ 0.53                 & 0.18                                    \\
	\multicolumn{1}{c}{}                              & \#M = 2           & 81.93 $\pm$ 6.63                & 77.22 $\pm$ 4.65                & 82.16 $\pm$ 6.48                & 76.23 $\pm$ 5.14               & 79.72 $\pm$ 6.00                & 74.41 $\pm$ 4.82                & 5.31                                    \\
	\multicolumn{1}{c}{}                              & \#M = 3           & 88.17 $\pm$ 3.21                & 85.00 $\pm$ 2.64                & 88.59 $\pm$ 2.70                & 85.69 $\pm$ 2.34               & 86.88 $\pm$ 5.51                & 84.47 $\pm$ 3.91                & 3.16                                    \\

	\multicolumn{1}{c}{}                              & \textit{Gau(0.1)} & 97.15 $\pm$ 0.04                & 96.38 $\pm$ 0.19                & 97.19 $\pm$ 0.05                & 96.41 $\pm$ 0.15               & 97.10 $\pm$ 0.07                & 96.39 $\pm$ 0.11                & 0.75                                    \\ 	
	
		\midrule
	%\cmidrule(r){1-1} \cmidrule(r){2-2} \cmidrule(r){3-3} \cmidrule(r){4-4} \cmidrule(r){5-5} \cmidrule(r){6-6} \cmidrule(r){7-7} \cmidrule(r){8-8} \cmidrule(l){9-9}
		
	\multicolumn{1}{l}{\multirow{5}{*}{Cifar10}}      & \textit{Dir(0.5)} & 95.08 $\pm$ 0.30                & 95.02 $\pm$ 0.37                & 95.74 $\pm$ 0.46                & 94.93 $\pm$ 0.30               & 94.95 $\pm$ 0.75                & 94.92 $\pm$ 0.41                & 0.30                                    \\
	\multicolumn{1}{c}{}                              & \#M = 1           & 16.42 $\pm$ 1.77                & 12.29 $\pm$ 0.84                & 16.65 $\pm$ 1.89                & 11.41 $\pm$ 1.33               & 17.23 $\pm$ 1.23                & 12.13 $\pm$ 0.67                & 4.81                                    \\
	\multicolumn{1}{c}{}                              & \#M = 2           & 75.66 $\pm$ 7.85                & 64.30 $\pm$ 6.08                & 75.53 $\pm$ 6.66                & 64.31 $\pm$ 5.31               & 74.43 $\pm$ 7.56                & 61.72 $\pm$ 7.93                & 11.76                                   \\
	\multicolumn{1}{c}{}                              & \#M = 3           & 87.32 $\pm$ 1.70                & 79.37 $\pm$ 1.43                & 87.29 $\pm$ 3.26                & 78.49 $\pm$ 1.92               & 85.03 $\pm$ 2.97                & 76.28 $\pm$ 1.89                & 8.50                                    \\

	\multicolumn{1}{c}{}                              & \textit{Gau(0.1)} & 94.68 $\pm$ 0.17                & 93.41 $\pm$ 0.27                & 94.90 $\pm$ 0.07                & 93.15 $\pm$ 0.27               & 94.59 $\pm$ 0.25                & 93.30 $\pm$ 0.26                & 1.44      \\
 
	\midrule
		%\cmidrule(r){1-1} \cmidrule(r){2-2} \cmidrule(r){3-3} \cmidrule(r){4-4} \cmidrule(r){5-5} \cmidrule(r){6-6} \cmidrule(r){7-7} \cmidrule(r){8-8} \cmidrule(l){9-9}

	\multicolumn{1}{l}{\multirow{5}{*}{FashionMNIST}} & \textit{Dir(0.5)} & 93.99 $\pm$ 0.64                & 92.82 $\pm$ 0.53                & 94.22 $\pm$ 0.54                & 93.14 $\pm$ 0.35               & 94.19 $\pm$ 0.52                & 93.12 $\pm$ 0.41                & 1.08                                    \\
	\multicolumn{1}{c}{}                              & \#M = 1           & 20.82 $\pm$ 2.00                & 16.47 $\pm$ 2.38                & 18.59 $\pm$ 2.12                & 17.92 $\pm$ 3.67               & 16.97 $\pm$ 1.16                & 17.03 $\pm$ 5.19                & 1.62                                    \\
	\multicolumn{1}{c}{}                              & \#M = 2           & 56.90 $\pm$ 5.26                & 55.40 $\pm$ 4.59                & 56.39 $\pm$ 3.09                & 55.62 $\pm$ 6.22               & 55.37 $\pm$ 3.20                & 52.83 $\pm$ 4.39                & 1.60                                    \\
	\multicolumn{1}{c}{}                              & \#M = 3           & 72.50 $\pm$ 3.89                & 70.90 $\pm$ 5.02                & 70.97 $\pm$ 6.94                & 67.94 $\pm$ 4.68               & 73.43 $\pm$ 7.46                & 67.31 $\pm$ 5.72                & 3.59                                    \\
	\multicolumn{1}{c}{}                              & \textit{Gau(0.1)} & 94.90 $\pm$ 0.12                & 93.52 $\pm$ 0.27                & 94.90 $\pm$ 0.13                & 93.56 $\pm$ 0.28               & 94.82 $\pm$ 0.07                & 93.52 $\pm$ 0.06                & 1.34                                    \\ 
	\midrule[1.3pt]

	\multicolumn{2}{l}{\textbf{Cross-Device}}                                      & \multicolumn{2}{c}{FedAvg}                                       & \multicolumn{2}{c}{FedProx}                                     & \multicolumn{2}{c}{FedNova}                                      &  \\ 
	
	\cmidrule(r){1-2} \cmidrule(r){3-4} \cmidrule(r){5-6} \cmidrule(r){7-8}
	
	\multicolumn{1}{c}{Dataset}                       & Partitioning      & \multicolumn{1}{l}{w/o adapter} & \multicolumn{1}{l}{w/ adapter} & \multicolumn{1}{l}{w/o adapter} & \multicolumn{1}{l}{w/adapter} & \multicolumn{1}{l}{w/o adapter} & \multicolumn{1}{l}{w/ adapter} & \multicolumn{1}{l}{Average difference}                   \\ 
	
	\cmidrule(r){1-1} \cmidrule(r){2-2} \cmidrule(r){3-3} \cmidrule(r){4-4} \cmidrule(r){5-5} \cmidrule(r){6-6} \cmidrule(r){7-7} \cmidrule(r){8-8} \cmidrule(){9-9}

	\multicolumn{1}{l}{\multirow{5}{*}{SVHN}}         & \textit{Dir(0.5)} & 95.64 $\pm$ 0.25                & 93.48 $\pm$ 1.42                & 95.62 $\pm$ 0.48                & 93.88 $\pm$ 0.90               & 77.86 $\pm$ 39.78               & 94.10 $\pm$ 0.90                & -5.26                                   \\
	\multicolumn{1}{c}{}                              & \#M = 1           & 14.70 $\pm$ 4.26                & 14.57 $\pm$ 4.03                & 14.49 $\pm$ 4.18                & 14.18 $\pm$ 4.63               & 11.19 $\pm$ 4.89                & 10.84 $\pm$ 5.23                & 2.89                                    \\
	\multicolumn{1}{c}{}                              & \#M = 2           & 89.59 $\pm$ 1.42                & 84.97 $\pm$ 0.91                & 89.99 $\pm$ 0.91                & 86.28 $\pm$ 4.04               & 49.81 $\pm$ 27.70               & 61.00 $\pm$ 25.27               & -0.72                                   \\
	\multicolumn{1}{c}{}                              & \#M = 3           & 94.16 $\pm$ 0.22                & 89.56 $\pm$ 0.78                & 94.05 $\pm$- 0.29               & 89.64 $\pm$ 0.35               & 92.60 $\pm$ 1.31                & 90.25 $\pm$ 1.50                & 2.24                                    \\

	\multicolumn{1}{c}{}                              & \textit{Gau(0.1)} & 96.78 $\pm$ 0,08                & 95.48 $\pm$ 0.13                & 95.58 $\pm$ 0.45                & 95.38 $\pm$ 0.14               & 96.81 $\pm$ 0.05                & 95.45 $\pm$ 0.08                & 1.29                                    \\ 
	
%	\cmidrule(r){1-1} \cmidrule(r){2-2} \cmidrule(r){3-3} \cmidrule(r){4-4} \cmidrule(r){5-5} \cmidrule(r){6-6} \cmidrule(r){7-7} \cmidrule(r){8-8} \cmidrule(l){9-9}
	\midrule

	\multicolumn{1}{l}{\multirow{5}{*}{Cifar10}}      & \textit{Dir(0.5)} & 94.78 $\pm$ 0.71                & 92.86 $\pm$0.98                 & 95.38 $\pm$ 0.79                & 92.91 $\pm$ 0.60               & 95.00 $\pm$ 0.93                & 92.63 $\pm$ 0.55                & 2.27                                    \\
	\multicolumn{1}{c}{}                              & \#M = 1           & 12.75 $\pm$ 1.55                & 10.41 $\pm$ 0.62                & 12.77 $\pm$ 2.31                & 10.66 $\pm$ 0.83               & 10.52 $\pm$ 1.07                & 9.97 $\pm$ 0.05                 & 12.43                                   \\
	\multicolumn{1}{c}{}                              & \#M = 2           & 89.64 $\pm$ 2.58                & 71.14 $\pm$ 3.94                & 89.76 $\pm$2.91                 & 71.50 $\pm$4.46                & 63.23 $\pm$ 4.09                & 58.99 $\pm$ 7.60                & 9.92                                    \\
	\multicolumn{1}{c}{}                              & \#M = 3           & 93.63 $\pm$ 0.49                & 80.92 $\pm$ 1.72                & 93.55 $\pm$ 0.65                & 80.74 $\pm$ 1.32               & 62.48 $\pm$ 29.87               & 82.61 $\pm$4.10                 & -5.90                                   \\

	\multicolumn{1}{c}{}                              & \textit{Gau(0.1)} & \textit{94.45 $\pm$ 0.23}       & 90.50 $\pm$ 1.12                & 94.42 $\pm$ 0.25                & 90.41 $\pm$ 0.85               & 94.32 $\pm$ 0.25                & 90.25 $\pm$ 1.05                & 4.00                                    \\ 
	
%	\cmidrule(r){1-1} \cmidrule(r){2-2} \cmidrule(r){3-3} \cmidrule(r){4-4} \cmidrule(r){5-5} \cmidrule(r){6-6} \cmidrule(r){7-7} \cmidrule(r){8-8} \cmidrule(l){9-9}
	\midrule
	
	\multicolumn{1}{l}{\multirow{5}{*}{FashionMNIST}} & \textit{Dir(0.5)} & 93.40 $\pm$ 0.62                & 92.19 $\pm$ 0.70                & 93.43 $\pm$ 0.81                & 92.22 $\pm$ 0.51               & 93.44 $\pm$ 0.32                & 92.13 $\pm$ 0.44                & 1.24                                    \\
	\multicolumn{1}{c}{}                              & \#M = 1           & 16.93 $\pm$ 3.47                & 11.32 $\pm$ 2.96                & 21.01 $\pm$ 3.21                & 10.52 $\pm$ 1.63               & 11.29 $\pm$ 1.73                & 11.16 $\pm$ 0.44                & 5.44                                    \\
	\multicolumn{1}{c}{}                              & \#M = 2           & 79.09 $\pm$ 5.32                & 72.85 $\pm$ 5.60                & 77.76 $\pm$ 4.87                & 76.76 $\pm$ 1.58               & 61.79 $\pm$ 28.07               & 47.76 $\pm$ 8.85                & 7.08                                    \\
	\multicolumn{1}{c}{}                              & \#M = 3           & 85.39 $\pm$ 3.21                & 85.42 $\pm$ 0.96                & 87.04 $\pm$ 2.20                & 85.41 $\pm$ 1.82               & 86.93 $\pm$ 1.97                & 81.94 $\pm$ 4.12                & 2.19                                    \\
	\multicolumn{1}{c}{}                              & \textit{Gau(0.1)} & \textit{94.29 $\pm$ 0.19}       & 92.77 $\pm$ 0.41                & 94.22 $\pm$ 0.14                & 92.48 $\pm$ 0.33               & 94.29 $\pm$ 0.14                & 92.55 $\pm$ 0.36                & 1.67                                    \\ 
	\bottomrule
\end{tabular}
}
\vspace{12pt}
\caption{Overview of resnet26 with and without adapters under Cross-Silo and Cross-Device distributions.} %various non-IID distributions}
\vspace{-10pt}
\label{tab:fullTableResults}
\end{table*}
The same tendencies for cross-silo are mostly true for cross-device, with the Adapters performing almost \emph{on par} with traditional finetuning of the entire model. The cross-device setting does show a more unstable training compared to cross-silo though. Interestingly adapters have lower variance in most cases for cross-silo settings, whilst the opposite is true in cross-device. 

On average we see a slightly better performance in the cross-silo setting.
% , however not by much. 
This may be the result of how well the participating clients $c_f$ data distribution represents the global distribution, as well as each client having on average a tenth of the data compared to our simulated cross-silo scenario. Since we only have a fraction of all clients in each round, this $c_{f}$ may not represent the global distribution. The optimization is therefore moving towards a local optimum of $c_{f}$ instead of an optima of the global problem. 
% It may be that for a lower $f$, the clients participating in a round poorly represent the global distribution of data and therefore move towards the local optimum of said clients instead of the optima of the global problem.

Both cross-silo and cross-device experiments show struggles with the quantity-based label imbalance experiments and especially with {M = 1}. This is true for all three FL algorithms. Fine tuning performs better than Adapters across the board for quantity-based label imbalance. A noteworthy observation is that quantity-based imbalance actually leads to better accuracy in the cross-device setting compared to cross-silo setting. 

%Across all three algorithms, we see comparable performance between Adapters and finetuning in the distribution-based label imbalance, especially so on Cifar10. Noise-based feature imbalance has more of a difference in cross-device settings, between Adapters and finetuning. This may be due to the larger spread of noise from the formula behind applying noise-based feature imbalance $Gau(\sigma * i/C)$, which is impacted by the number of clients. 

\begin{figure}[]
   \centering
   \includegraphics[width=.98\linewidth]{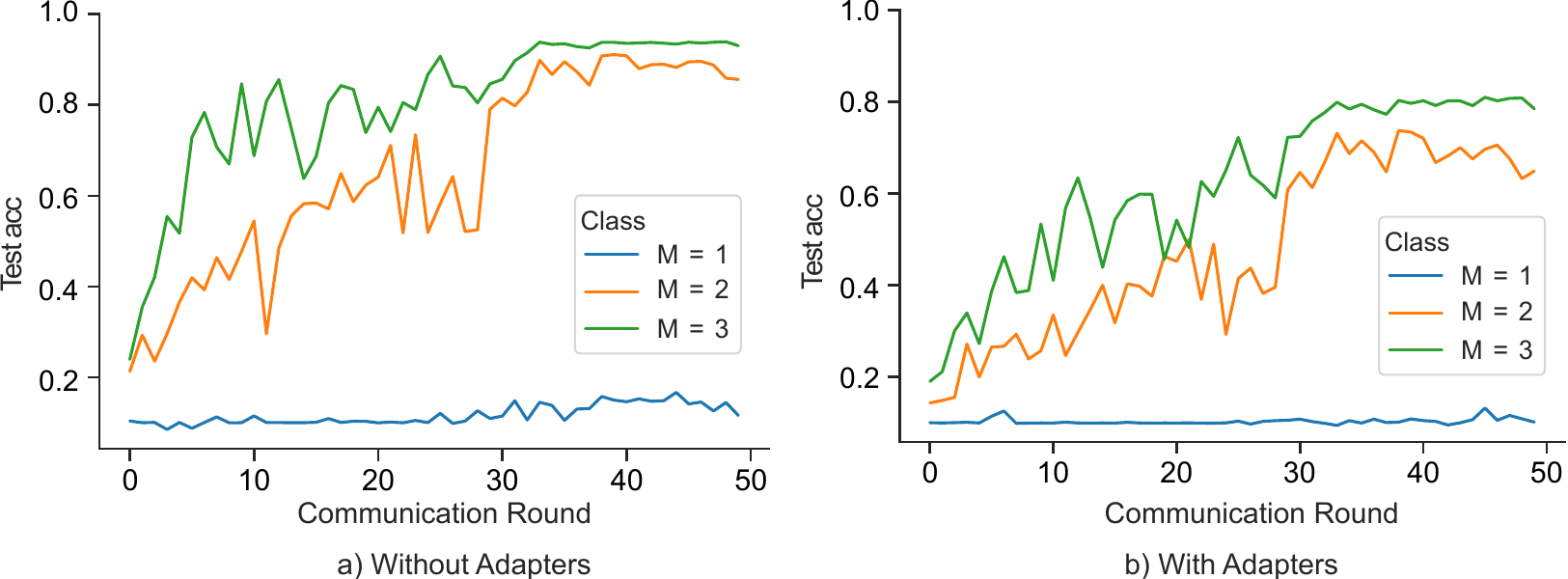}%fedpCifarClassVariance.png}
    \caption{FedProx with \textit{mu} $0.01$ on Cifar10, under different Quantity-based label imbalances, ranging from $M = 1, 2, 3$ in the cross-device setting. Learning rate is $0.01$}
    \vspace{-10pt}
    \label{fig:ClassCifar10Fedp}
\end{figure}

\textbf{Instabilities during training} were seen in a few cases. Batch normalization under non-IID data distributions may introduce instability during the training phase~\cite{DirichletFLSilo, fedbn}. Batch normalization standardizes inputs to each layer and records statistics of the data distribution the layer observes. Since we divide the data on clients based on heterogeneous distributions, each client has different statistics. The averaging step of FL algorithms leads to batch normalization layers that may not represent the local data distributions during training nor the global test set completely. It was proposed in~\cite{fedbn} to keep the running mean and variance of the batch normalization layers local for each client. %We do not use this approach, but it may 
This might be beneficial for further research when handling certain non-IID cases. 
When distributing the data based on quantity-based label imbalance, the stability of the training depends on the amount of classes of each client. 
% the lower amount of classes each client has, the more unstable the training becomes. 
For $M=1$, we see almost no progress in the training of either Adapters or the full model. Previous work observed unstable training when distributing based on the quantity-based label imbalance with a low number of classes $M$ even without batch normalization~\cite{DirichletFLSilo}. This instability was even more pronounced with a higher LR, and therefore we lowered the LR for quantity-based label imbalance experiments. Figure \ref{fig:ClassCifar10Fedp} shows the training under label-based label imbalance with and without Adapters using FedProx with $M = 1$, $2$ and $3$ with LR of $0.01$. For $M = 1$, the model is barely able to learn anything. For $M = 2$ the model starts to learn, but has worse performance and stability than $M = 3$. 

%This instabillity was even more pronounced with a higher learning rate, and was why we lowered the lr for all label-based label i
%
%We lower the learning rate and see the instabillity reduces accordingly, this is illustrated on Figure \ref{fig:lrInstabillity}, and is why we use a lower lr $0.01$ on the experiments with quantity-based label imbalances. An important note is that the instabillity is seen for both adapters and finetuning. 

%\begin{figure}[!h] 
%    \centering
%  \subfloat[M = 1\label{1a}]{%
%       \includegraphics[width=0.33\linewidth]{images/Fig_6/fedAvg_svhn_LrC1_1.pdf}}
%    %
%  \subfloat[M = 2\label{1b}]{%
%        \includegraphics[width=0.33\linewidth]{images/Fig_6/fedAvg_svhn_LrC2_1.pdf}}
        %
%  \subfloat[M = 3\label{1c}]{%
%        \includegraphics[width=0.34\linewidth]{images/Fig_6/fedAvg_svhn_LrC3_1.pdf}}
%\newline
%  \caption{Fedavg on quantity-based label imbalances showing the instability of training with different learning rates on SVHN}
%  \label{fig:lrInstabillity} 
%\end{figure}

FedNova as seen on Figure \ref{fig:fedAlgsCrossSilo} suffers from a more pronounced unstable training phase. We believe this to be from the normalized gradients to the batch normalization parameters sometimes result in invalid values (i.e., NaN) as the output of the global aggregated model. This is reflected as abrupt drops in the aggregated performance. A lower LR does resolve most of these issues in distribution-based label imbalance tests. However the final test performance, when comparing the two learning rates, is higher with an initial LR of $0.1$ for both Adapters and finetuning.

With Adapters we can reduce the amount of parameters that needs to be exchanged by a factor of approximately 9. We compare the data exchanged between the client and the server and measure the fully finetuned model against the amount of data required when only exchanging the Adapters and domain specific information.
We compare the transmitted data size of the finetuned model and the Adapters by saving the state\_dicts of each transmission. The finetuning approach sends all parameters of the Resnet26, for which the payload size is 22.2 MB. With Adapters the full model grows to 24.7 MB. We remove the frozen weights of the base model from the state\_dict and it is reduced to a size of 2.58 MB. This means we can effectively lower the payload of each FL round by almost a ninth. 
%
%The first round of communication has to distribute the base model as well, unless this is done in some other secure fashion. 

% If a client in FL has multiple tasks for which a DNN is trained, with Adapters we can use the same base model and train domain specific Adapters for each. This means, for each task trained we can reduce the required storage, since we only need the domain specific Adapters~\cite{AdaptersDomain}. 
If a client in FL has multiple tasks for which a DNN is trained, we can use the same base model and train domain specific Adapters for each domain. This means, for each task trained we can reduce the required storage, since we only need the domain specific Adapters~\cite{AdaptersDomain}. 

\section{Conclusion}
\label{sec:conclusion}
In this paper we proposed the use of Adapters in Federated Learning to reduce communication cost in each round between server and clients. We showed how Adapters can be used as an addition to existing FL algorithms. Furthermore we conducted comprehensive experiments comparing Adapters to finetuning through multiple FL algorithms, datasets, and non-IID strategies. Adapters perform \emph{on par} or close for both quantity-based label imbalance and noise-based feature distribution skew. For distribution-based label imbalance we see a more pronounced difference between finetuning and Adapters indicating a vulnerability of the Adapter-based FL approach. Clients in FL can further benefit from Adapters if they are subject to multi-domain tasks, where Adapters can lead to reduced storage of models, since the domain agnostic base model can be used for multiple domains. 

There are several directions for future work. One avenue to explore is the potential of Adapters to further improve the existing security and privacy benefits of FL. If one can distribute the parameters of the base model in a secure fashion, this can further increase security of the final network, since the use of Adapters requires the base network parameters to function. Therefore, if one has access to the Adapters in a transmission round, he/she also needs the base network parameters to be able to use the model. Adapters are initialized by the coordinating party, to ensure the same starting point for all involved parties.

%\section*{Acknowledgement}
%\hl{Removed for Double blind}

% The very first letter is a 2 line initial drop letter followed
% by the rest of the first word in caps.

% trigger a \newpage just before the given reference
% number - used to balance the columns on the last page
% adjust value as needed - may need to be readjusted if
% the document is modified later
%\IEEEtriggeratref{8}
% The "triggered" command can be changed if desired:
%\IEEEtriggercmd{\enlargethispage{-5in}}

% references section

% can use a bibliography generated by BibTeX as a .bbl file
% BibTeX documentation can be easily obtained at:
% http://mirror.ctan.org/biblio/bibtex/contrib/doc/
% The IEEEtran BibTeX style support page is at:
% http://www.michaelshell.org/tex/ieeetran/bibtex/
 \bibliographystyle{IEEEtran}
 \bibliography{references.bib}

\end{document}